\documentclass{article}

\usepackage{gen2_iclr2026tiny,times}
\iclrfinalcopy

\usepackage{amsmath,amsfonts,bm}

\def\eqref#1{equation~\ref{#1}}

\def\1{\bm{1}}

\DeclareMathAlphabet{\mathsfit}{\encodingdefault}{\sfdefault}{m}{sl}
\SetMathAlphabet{\mathsfit}{bold}{\encodingdefault}{\sfdefault}{bx}{n}

\usepackage{hyperref}
\usepackage{url}
\usepackage{algorithm}
\usepackage{algorithmicx}
\usepackage{algpseudocode}
\usepackage{float}
\usepackage{graphicx}

\title{Continuous Diffusion Transformers for Designing Synthetic Regulatory Elements}

\author{Jonathan Liu$^*$ \& Kia Ghods\thanks{ Equal Contribution} \\
Department of Computer Science\\
Princeton University\\
\texttt{\{jonathanliu,kia.ghods\}@princeton.edu} \\
}

\begin{document}

\maketitle

\begin{abstract}
\vspace{-1em}
We present a parameter-efficient Diffusion Transformer (DiT) for generating 200\,bp cell-type-specific regulatory DNA
  sequences. By replacing the U-Net backbone of DNA-Diffusion \citep{dasilva2025designing} with a transformer denoiser equipped with a 2D CNN input encoder, 
  our model matches the U-Net's best validation loss in 13 epochs (60$\times$ fewer) and converges 39\% lower, while reducing
  memorization from 5.3\% to 1.7\% of generated sequences aligning to training data via BLAT. Ablations show the CNN encoder is
  essential: without it, validation loss increases 70\% regardless of positional embedding choice. We further apply DDPO
  finetuning using Enformer as a reward model, achieving a 38$\times$ improvement in predicted regulatory activity.
  Cross-validation against DRAKES on an independent prediction task confirms that improvements reflect genuine regulatory signal
   rather than reward model overfitting.
\end{abstract}
\vspace{-1em}

\section{Introduction}
The ability to generate short DNA sequences with designated regulatory effects remains a bottleneck for safe and precise genetic modulation. Existing approaches broadly fall into (1) DNA foundation models and (2) small-insert, objective-driven generators (often diffusion-based). While transformers have achieved strong performance in sequence modeling, conditioning and controllability for regulatory design remain challenging. 

Specifically, we use a Diffusion Transformer to learn the diffusion process \citep{peebles2023scalable}. We choose a transformer model because U-nets---which have fixed receptive fields---fail to model long-distance DNA interactions. Works in the literature have not trained diffusion models to generate short-inserts that optimize regulatory activity, however, our lightweight models allow us to conduct normally expensive rollouts during RL finetuning. As a result, our final model is capable of returning DNA segments that natively have high predicted promoter activity and DNA accessibility.

We make the following contributions: (1) \textbf{Continuous DiT for regulatory design:} We develop a parameter-efficient transformer-based diffusion model for generating synthetic 200 bp regulatory elements under cell-type-specific objectives that surpasses the performance of previous models in 60x fewer steps and with 6x fewer parameters; and (2)
\textbf{Post-training with RL optimization:} We perform RLVR-style finetuning using Enformer as a reward model to improve accessibility/activity proxies. Using DRAKES as a verifier, we find that our RL provides signal to similar tasks \citep{wang2024fine}.

\section{Related Works}

Deep learning for regulatory genomics advanced through \textit{sequence-to-function} predictors such as DeepSEA \citep{zhou2015deepsea}, Basset \citep{kelley2016basset}, Basenji \citep{kelley2018basenji}, BPNet \citep{avsec2021bpnet}, and Enformer \citep{avsec2021effective}. We use Enformer (cell-type-specific CAGE/DNase from 196kb context) as an evaluation oracle. These predictors naturally induce an inverse problem: can we \textit{design} sequences that achieve desired regulatory behaviors?

Generative approaches fall into two regimes: (1) large \textit{DNA foundation models} (autoregressive or masked) that capture broad genomic structure, and (2) \textit{small-insert, objective-driven generators} that produce short cis-regulatory candidates conditioned on cell type and assay objectives. DNA-Diffusion \citep{dasilva2025designing} demonstrated that diffusion models in the second regime can propose diverse, motif-plausible candidates scoring highly under chromatin predictors, using a U-Net denoiser. Our work replaces this with a parameter-efficient DiT backbone and adds post-training alignment to predictor objectives.

\section{Methods}
\subsection{Data}
We consider the problem of generating synthetic 200 bp DNA sequences intended to function as regulatory elements that increase cell-type-specific activity under learned proxy predictors following \cite{dasilva2025designing}. Specifically, we focus on designing regulatory elements that enhance the gene activity in K562, HepG2, GM12878, and hECT0 cells. Starting with ENCODE DHS data which identify DNase I hypersensitive sites (DHSs) that mark regions of open chromatin and regulatory activity, we identify peaks of DNAseI activity to define cis-regulatory regions in each cell line. Following DNA-Diffusion, we use a dataset with 12k samples from each of the 4 cell lines, ensuring that the sequences of DNA do not repeat between the cells. In total, our dataset consists of 47,872 sequences.

\subsection{Model formulation and preliminaries}
We train a diffusion model to denoise corrupted continuous representations of DNA sequences following the standard DDPM protocol from \cite{ho2020denoising}. Specifically, we use Adam~\citep{kingma2017adam} with learning rate $2 \times 10^{-4}$, bf16 mixed precision, and batch size $1024$. The diffusion process uses $100$ timesteps with a linear noise schedule from $\beta_{\text{start}}=0.296$ to $\beta_{\text{end}} = 0.25$ and unconditional dropout $p_{\mathrm{uncond}} = 0.1$ for classifier-free guidance. 

Our DiT (dim$=320$, depth$=6$, $8$ heads) uses AdaLN-Zero conditioning with learned positional embeddings. The $4 \times 200$ one-hot input is processed through a 2D CNN encoder (kernel size $5$) that treats the nucleotide$\times$position matrix as a spatial feature map, capturing local k-mer structure before the transformer layers. Architecture and alternative input format ablations are reported in \autoref{sec: ablation}. The U-Net baseline follows the original DNA-Diffusion architecture~\citep{dasilva2025designing} with $\mathrm{dim}=200$ and channel multipliers $[1,2,4]$. We train for a minimum of $2{,}000$ epochs with early stopping (patience $10$). At inference, we use classifier-free guidance with scale $w=2.0$.

\subsection{Post-training via reinforcement learning / predictor-guided finetuning}
Our finetuning setup is RL-algorithm agnostic and simply modifies the inference setup and reward signal. At each training iteration, we randomly sample a target cell type and condition both the diffusion sampler and reward function on this cell. Candidate sequences are generated via classifier-free guidance using the selected cell embedding, and rewards are computed from the corresponding Enformer output track predicting the CAGE for that cell \citep{avsec2021effective}. Policy updates are then performed using denoising diffusion policy optimization (DDPO) \citep{black2023training}. The training loop is detailed in Appendix \autoref{appendix:rl}.

We consider two training scenarios. In the \textit{in situ} setting, Enformer evaluates the generated DNA embedded within the GATA1 locus, testing the model’s ability to produce sequences that interact with distal genomic context. In the \textit{ex situ} setting, the enhancer is evaluated in isolation, requiring the model to encode enhancer-specific structure directly within the 200 bp insert.

\subsection{Cross-Validation} Finally, we verify that our model does not overfit to the Enformer model. Using an oracle for predicting HepG2 activity, we evaluate our conditional DNA generation compared to that of single-cell diffusion model DRAKES \citep{wang2024fine}. Notably, our dataset (with 12k HepG2 sequences) is unlabeled, whereas the DRAKES model utilizes a 700k dataset of Enhancers. Due to the difference in training task and data, our goal in this comparison is not to match absolute performance, but rather to confirm that our generations exhibit a meaningful structural signal that generalizes to these constraints.

\section{Experiments}
\subsection{Generation quality}

\begin{figure}[H]
    \centering
    \includegraphics[width=.75\linewidth]{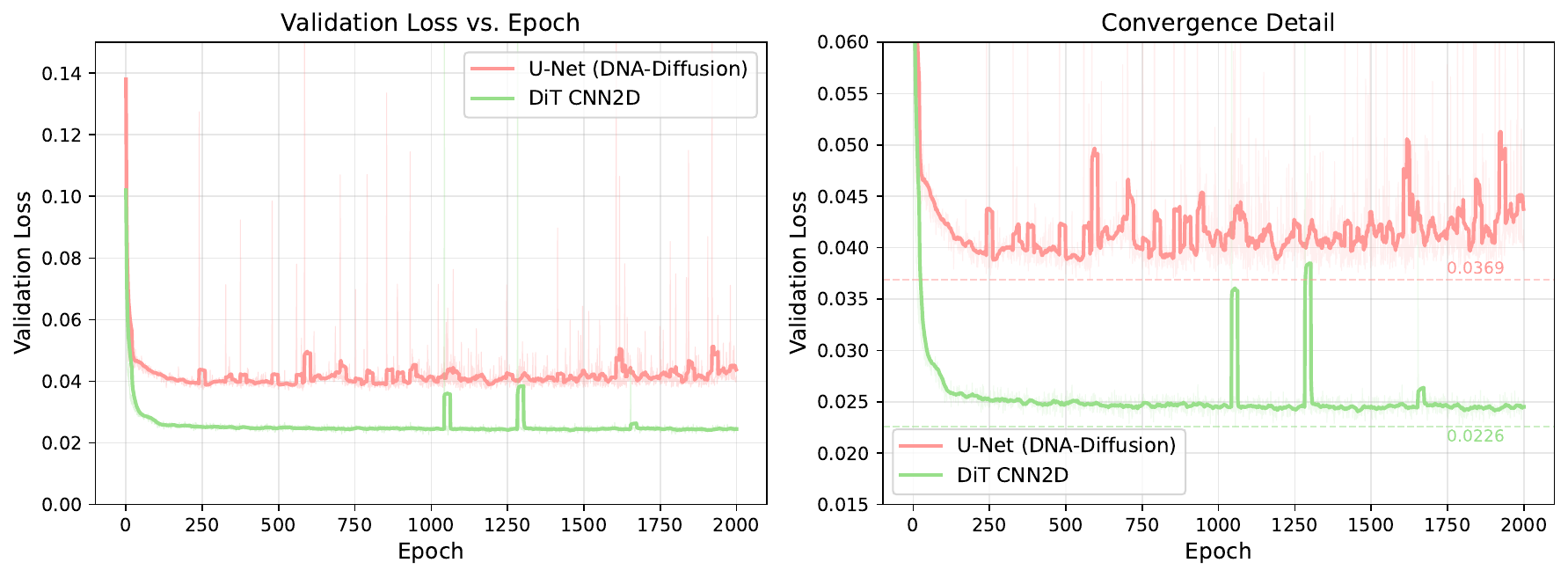}
    \caption{Loss Curve Comparison of the U-Net and our DiT. }
    \label{fig:loss}
    \vspace{-1em}
\end{figure}
Our DiT matches the best validation loss of the U-Net baseline within 13 epochs ($\sim$60$\times$ fewer), and ultimately converges 39\%
lower (0.023 vs.\ 0.037), as illustrated in \autoref{fig:loss}.

We note that a key concern with generative DNA models is memorization, in this context meaning that the model produces near-copies of training sequences rather than novel regulatory candidates. We evaluate this with two complementary analyses. BLAT alignment \citep{kent2002blat} queries each generated sequence against the full training set for high-identity matches ($\geq$20\,bp, $\geq90$\% identity); a high match rate indicates memorization. Motif JS distance scans all sequences for 879 transcription factor (TF) binding motifs from JASPAR \citep{rauluseviciute2024jaspar} and compares their frequency distribution s between generated and held-out test sequences via Jensen--Shannon divergence; low JS distance indicates that the model has learned biologically realistic motif usage rather than copying specific sequences. 

\begin{figure}[H]
    \centering
    \includegraphics[width=.75\linewidth]{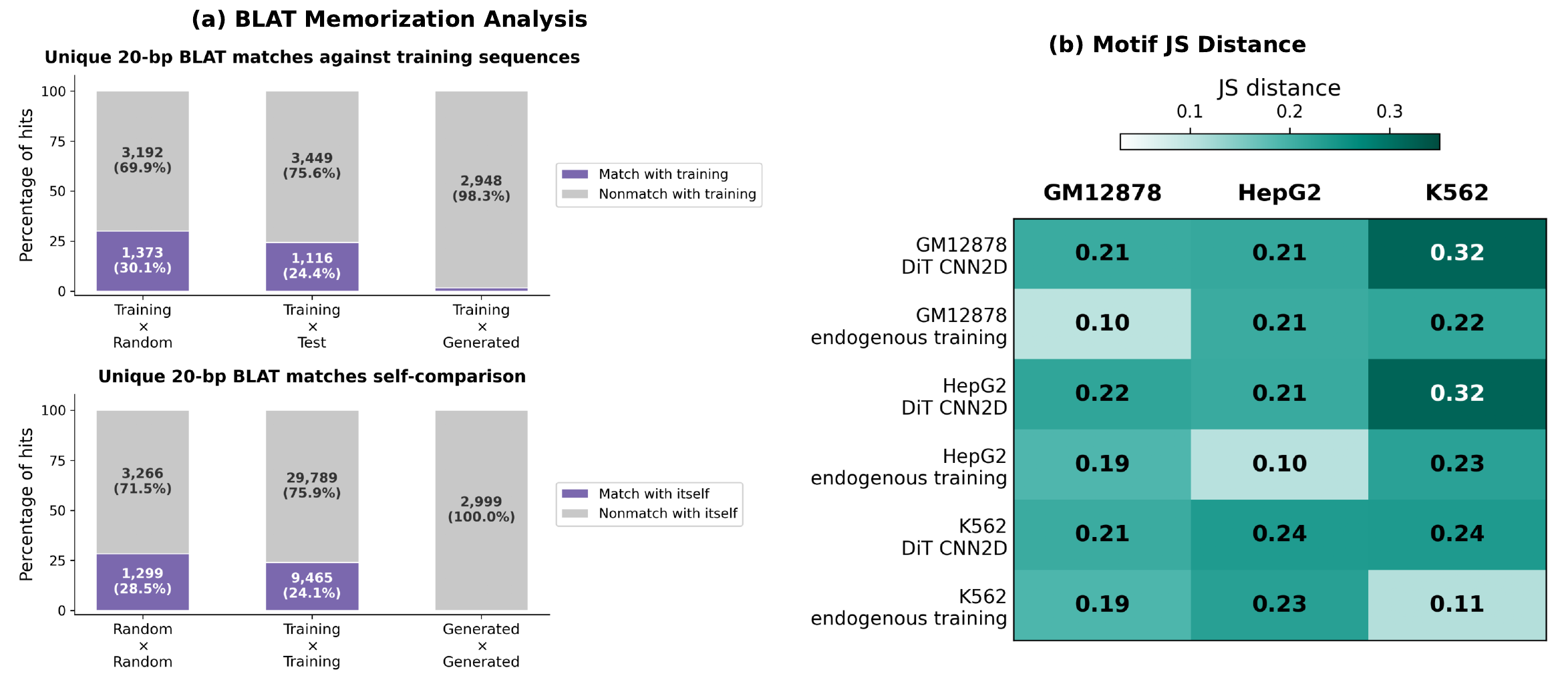}
    \caption{Memorization and Modeling Analysis. (a) Blat Memorization Analysis counting the unique 20-bp BLAT matches across Training, Test, Generated, and Random sequences of DNA. (b) JS Distance comparing the distances between distributions of DNA in our generated DNA and the endogenous DNA sequences.}
    \label{fig:2ab}
    \vspace{-1em}
\end{figure}
We demonstrate in \autoref{fig:2ab} that the generation quality of our DiT matches prior DNA-Diffusion baselines on motif recovery (JS
distance) while exhibiting substantially less memorization: only
1.7\% of generated sequences align to the training set via BLAT \citep{kent2002blat},
compared to 5.3\% for the U-Net reported
by~\citet{dasilva2025designing}. We attribute this to the
transformer's global attention mechanism, which avoids the fixed
receptive fields of convolutional architectures.

\subsection{Reinforcement Learning}
Utilizing our training methodology and DDPO, we find that we are able to increase the predicted in-situ expression over 38x compared to the baseline model on average \autoref{tab:model_comparison}. Evaluation was conducted across 250 sequences generated for each cell line. We present the best RL results here, though a sweep of RL hyperparameters can be found in \ref{appendix:rl-hyp}. In \autoref{fig:In-Situ_enformer}, we additionally see that over 75 percent of all generations, across all cell types, have higher Enformer score than the baseline median. 

\subsection{Cross-Validation}
For validation, we compare our model to the reported values of the DRAKES model, which is optimized for maximizing single-cell (HepG2) expression. Our model captures 70\% (3.86) of the 5.6 predicted activity by DRAKES, suggesting the presence of a meaningful signal under these constraints. 

\begin{table}[t]
\centering
\begin{tabular}{llcccc}
Model & Mode & GM12878 & HepG2 & K562 & hESCT0 (DNAse)\\
\hline
DNA-Diffusion & In-Situ & 0.53399 & 0.06957 & 0.05587 & -- \\
\hline
CNN-DiT & In-Situ & 0.19025 & 0.04854 & 0.59428 & 0.20012 \\
CNN-DiT & Ex-Situ & 0.07316 & 0.04656 & 0.03551 & 0.03540 \\
\hline
CNN-DiT-DDPO & In-Situ & \textbf{4.19501} & \textbf{4.11424} & \textbf{4.76197} & 1.86090 \\
CNN-DiT-DDPO & Ex-Situ & \textbf{1.16401} & \textbf{1.17669} & \textbf{0.40109} & 0.28727 \\
\hline
\end{tabular}
\caption{Cell-type-specific median activity scores predicted by Enformer of generated 200 bp regulatory sequences across models and training modes. The hESCT0 predictions use the DNAse activity. Bolded values are best for their category. In-situ denotes the predictions when the 200bp sequence is embedded into the GATA1 enhancer DNA. The Ex-Situ condition predicts the activity score using only the 200bp sequence surrounded by filler tokens ($[0.25, 0.25, 0.25, 0.25]$)}.
\label{tab:model_comparison}
\end{table}

\begin{figure}
    \centering
    \includegraphics[width=0.7\linewidth]{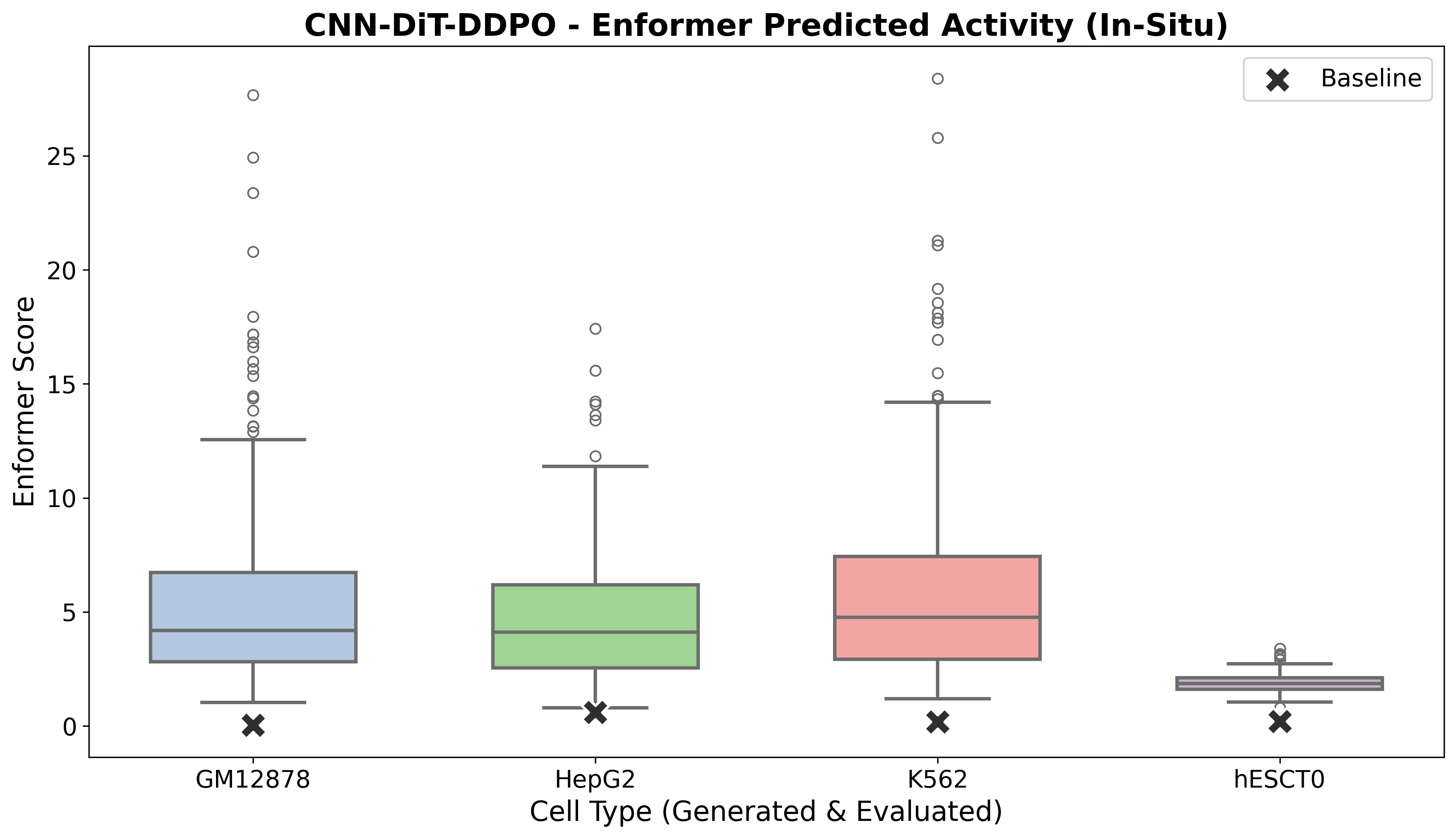}
    \caption{Distribution of Enformer-predicted In-Situ Predicted Activity from 250 generated sequences. Black crosses denote the median predictions of the pre-trained model.}
    \label{fig:In-Situ_enformer}
    \vspace{-1em}
\end{figure}

\subsection{Alternative Modeling Approaches}
\label{sec: ablation}

\paragraph{Positional embeddings without CNN input preprocessing.}
We trained DiT variants using a linear projection of the $4 \times 200$ one-hot input (no convolutional encoder) with either RoPE or learned positional embeddings. Both converged to validation losses of 0.038--0.039, roughly 70\% higher than the CNN2D model's 0.023, with the gap persisting across 2{,}000 epochs.

Interestingly, RoPE achieved comparable motif JS distances to CNN2D (between 
0.20--0.21 vs.\ 0.21--0.22), while learned embeddings fared worse (0.24--0.25). RoPE's relative-position awareness partially compensates for lacking local inductive bias, but cannot fully replace the CNN encoder's k-mer feature extraction. This is consistent with the broader observation that transformers benefit from convolutional stems for spatially structured inputs.

\section{Discussion and Limitations}
Our results demonstrate that transformer-based diffusion models can substantially outperform U-Net baselines for regulatory DNA generation, but only when equipped with appropriate inductive biases. As our ablation (\autoref{sec: ablation}) shows, the CNN encoder is essential: without it, validation loss increases 70\% regardless of positional embedding choice, confirming that transformers require convolutional stems to capture local structure in spatially organized inputs.

The 38$\times$ improvement in predicted expression from DDPO finetuning is encouraging, but carries important caveats. Indeed, post-hoc analysis of DDPO-finetuned generations reveals a distribution shift: while memorization of training data remains low (3.0\% BLAT), self-alignment rises to 92.8\% indicating that the policy converges to a narrow distribution. Enformer, while state-of-the-art, is an imperfect proxy: optimized sequences may exploit model-specific biases rather than genuine regulatory logic. Our DRAKES comparison partially mitigates this concern by showing that improvements transfer to an independent predictor and task, but the use of other validation models (BORZOI or AlphaGenome) and wet-lab validation (e.g., MPRA assays) remains necessary to confirm functional activity. Additionally, our 200\,bp generation window cannot capture distal regulatory interactions, and our balanced 12k-per-cell-type dataset is small relative to the full ENCODE dataset. Future work will explore scaling to longer inserts, larger multi-cell datasets, and closed-loop experimental validation.
\bibliography{gen2_iclr2026_workshop}
\bibliographystyle{gen2_iclr2026_workshop}
\appendix
\section{Appendix}
\subsection{Dataset}
\begin{table}[H]
\centering
\begin{tabular}{ll}
\hline
\textbf{Cell Line} & \textbf{ENCODE Accession} \\ \hline
hESCT0           & ENCLB449ZZZ \\
K562             & ENCLB843GMH \\
HepG2            & ENCLB029COU \\
GM12878          & ENCLB441ZZZ \\
\hline
\end{tabular}
\caption{Summary of Cell Lines and ENCODE Accession Numbers}
\label{dataset_details}
\end{table}

\subsection{DiT Hyperparameter sweep}

\label{sec:sweep}
We conducted a hyperparameter sweep over 96 trials (83 completed, 13 pruned) using Optuna \citep{akiba2019optunanextgenerationhyperparameteroptimization} with TPE sampling, comparing DiT and UNet architectures on the 48k dataset. Each trial trained for up to 3{,}000 epochs with early stopping (patience 10). The DiT search space covered: dim $\in \{192, 256, 320, 384\}$, depth $\in \{6, 8, 10, 12\}$, heads $\in \{6, 8, 12\}$, dim\_head $\in \{48, 64, 80\}$, MLP ratio $\in \{3, 4, 5\}$, dropout $\in \{0, 0.02, 0.05\}$, timesteps $\in \{50, 75, 100\}$, $\beta_{\mathrm{end}} \in \{0.15, 0.2, 0.25\}$, $\beta_{\mathrm{start}} \in [5\mathrm{e}{-5}, 5\mathrm{e}{-4}]$, lr $\in [3\mathrm{e}{-5}, 3\mathrm{e}{-4}]$, and batch size $\in \{512, 768, 1024, 1280\}$.

\paragraph{Key findings.} The top 5 trials all converged to the same configuration: dim$=$320, depth$=$6, heads$=$8, dim\_head$=$48, mlp\_ratio$=$5.0, timesteps$=$100, $\beta_{\mathrm{end}}=0.25$, learned\_sinusoidal\_dim$=$32 (val\_loss $\in [0.0219, 0.0225]$). The only variation among top trials was in learning rate ($1.8$--$2.0 \times 10^{-4}$), $\beta_{\mathrm{start}}$ ($2.3$--$3.1 \times 10^{-4}$), dropout ($0.0$ vs $0.02$), and batch size ($1024$ vs $1280$). This suggests the architecture is robust to minor hyperparameter variation once the structural choices are fixed.

In contrast, the worst-performing trials (val\_loss $> 0.04$) were characterized by deeper models (depth$=$8--10), smaller dimensions (dim$=$192--384), fewer timesteps ($50$--$75$), and lower $\beta_{\mathrm{end}}$ ($0.15$--$0.2$). The clearest negative signals were: (1) depth $> 6$ consistently hurt, suggesting overfitting for this dataset size; (2) timesteps $< 100$ degraded performance; and (3) lower $\beta_{\mathrm{end}}$ values produced insufficient noise corruption.

The best trial configuration (Trial 14, val\_loss$=$0.0219) was adopted as the final DiT architecture for all subsequent experiments, with the 2D CNN input encoder added post-sweep.

\subsection{RL Algorithm Specifics}
\label{appendix:rl}
\begin{algorithm}[H]
\begin{algorithmic}[1]
\Require Set of cell types $\mathcal{C}$, diffusion policy $\pi_\theta$, frozen proxy predictor $f$ (Enformer)
\State Initialize diffusion policy parameters $\theta$
\State Freeze proxy predictor $f$
\For{each training iteration}
    \State Sample cell type $c \sim \mathrm{Uniform}(\mathcal{C})$
    \State Generate sequence $x \sim \pi_\theta(\cdot \mid c)$ using classifier-free guidance
    \State Compute reward $r \gets f_c(x)$
    \State Update $\theta$ using the RL objective (DDPO)
\EndFor
\end{algorithmic}
\caption{Task-Conditioned Diffusion RL for Cell-Specific Regulatory Design}
\label{alg:cell_cond_diff_rl}
\end{algorithm}

\subsection{RL Algorithm Hyperparameters}
\label{appendix:rl-hyp}
The DDPO algorithm conducts finetuning efforts swept over two hyperparameters: $\text{lr} \in \{1e-5, 5e-5\}$ and $\text{ppo\_epochs} \in  \{4, 8, 16, 32\}.$ Each run trained for 5000 steps with a batch size of $16$, $\beta=0.5.$ We found that the best configuration was $\text{lr}=5e-5$ and $\text{ppo\_epochs}=4.$ We additionally considered SDPO and GRPO though we found early on that their performance was much worse than that of DDPO.

\subsection{Model configurations}
\label{sec:model_configs}

\begin{table}[H]
\centering
\small
\begin{tabular}{lcc}
\hline
\textbf{Parameter} & \textbf{U-Net} & \textbf{DiT CNN2D} \\
\hline
Backbone & 2D Conv & Transformer \\
Input encoding & Direct $4 \times 200$ & CNN2D (5x4 kernel) \\
Positional embedding & -- & Learned \\
Hidden dim & 200 & 320 \\
Depth / Layers & 3 (mults $[1,2,4]$) & 6 \\
Attention heads & 4 & 8 (dim\_head$=$48) \\
MLP ratio & -- & 5.0 \\
Conditioning & Additive & AdaLN-Zero \\
Dropout & 0.0 & 0.02 \\
Diffusion timesteps & 50 & 100 \\
Noise schedule & Linear & Linear \\
$\beta_{\mathrm{start}}$ & $1 \times 10^{-4}$ & $3 \times 10^{-4}$ \\
$\beta_{\mathrm{end}}$ & 0.2 & 0.25 \\
$p_{\mathrm{uncond}}$ (CFG) & 0.1 & 0.1 \\
Batch size & -- & 1024 \\
Learning rate & -- & $2 \times 10^{-4}$ \\
RC augmentation & No & Yes (50\%) \\
Best val loss & 0.037 & \textbf{0.023} \\
\hline
\end{tabular}
\caption{Architecture and training configuration comparison between the U-Net baseline (DNA-Diffusion) and our DiT CNN2D model. U-Net values are from the pretrained checkpoint; entries marked ``--'' were not reported.}
\label{tab:model_configs}
\end{table}

\end{document}